\documentclass[journal]{IEEEtran}

\usepackage{cite}

\ifCLASSINFOpdf
  \usepackage[pdftex]{graphicx}
  \graphicspath{{../pdf/}{../jpeg/}}
  \DeclareGraphicsExtensions{.pdf,.jpeg,.png}
\else
  \usepackage[dvips]{graphicx}
  \graphicspath{{../eps/}}
  \DeclareGraphicsExtensions{.eps}
\fi

\usepackage[cmex10]{amsmath}
\usepackage{amssymb}

\begin{document}
\title{Optimal Discriminant Functions Based On Sampled Distribution Distance for Modulation Classification}
\author{Paulo~Urriza,~\IEEEmembership{Student Member,~IEEE,}
        Eric~Rebeiz,~\IEEEmembership{Student Member,~IEEE,}
        and~Danijela~Cabric,~\IEEEmembership{Member,~IEEE}
\thanks{The authors are with the Department of Electrical Engineering, University
of California, Los Angeles, 56-125B Engineering IV Building, Los
Angeles, CA 90095-1594, USA (e-mail: \{pmurriza, rebeiz, danijela\}@
ee.ucla.edu).}
\thanks{This work is supported by DARPA under grant A002069701.}
} 

\maketitle

\begin{abstract}
In this letter, we derive the optimal discriminant functions for modulation classification based on the sampled distribution distance. The proposed method classifies various candidate constellations using a low complexity approach based on the distribution distance at specific testpoints along the cumulative distribution function. This method, based on the Bayesian decision criteria, asymptotically provides the minimum classification error possible given a set of testpoints. Testpoint locations are also optimized to improve classification performance. The method provides significant gains over existing approaches that also use the distribution of the signal features.
\end{abstract}

\IEEEpeerreviewmaketitle


\section{Introduction}
Modulation classification is the process of choosing the most likely scheme from a set of predefined candidate schemes that a received signal could belong to. Various approaches have been proposed to address this problem. There has recently been growing interest in modulation classification for applications such as software defined radio, cognitive radio and interference identification \cite{Lee2011}.

Existing classification methods can generally be categorized into two main groups: feature based classifiers and likelihood based (ML) classifiers. The ML classifiers give the minimum possible classification error of all possible discriminant functions given perfect knowledge of the signal's probability distribution. However, this approach is very sensitive to modeling errors such as imperfect knowledge of the signal to noise ratio (SNR) or phase offset. Further, such approaches have very high computational complexity and are thus impractical in actual hardware implementation. To address these issue, various feature based techniques such as cumulant-based classifiers \cite{swami_tcom_2002} and cylostationary-based classifiers have been proposed \cite{Rebeiz2011a}.

Recently, Goodness-of-Fit (GoF) tests such as the Kolmogorov-Smirnov (KS) \cite{Stephens1974} distribution distance have been proposed to identify the constellation used in QAM modulation \cite{Wang2010}. Based on the KS classifier, we proposed a new reduced complexity Kuiper (rcK) classifer in \cite{Urriza2011}. The rcK classifier only finds the empirical cumulative distribution function (ECDF) in a small set of predetermined testpoints that have the highest probability of giving the maximum distribution distance, effectively sampling the distribution function. The algorithm offered reduced computational complexity by removing the need to estimate the full ECDF while still providing better performance than the KS classifier. It also increased the robustness of the classifier to imperfect parameter estimates.

The idea of improving the classification accuracy of the rcK classifier by using more testpoints was proposed in \cite{Wang2012}. The method is referred to as Variational Distance (VD) classifier where testpoints are selected to be the pdf-crossings of two classes being recognized. The sum of the absolute distances is then used as the final discriminating statistic. We refer to methods such as rcK and VD, that utilize the value of the ECDF at a small number of testpoints, as sampled distribution distance classifiers. In this work we derive the optimal discriminant functions for classification with the sampled distribution distance given a set of testpoint locations. We also provide a systematic way of finding testpoint locations that provide near optimal performance by maximizing the Bhattacharyya distance between classes. Finally, we present results that compare the performance of this approach with existing techniques.

\section{Proposed Classifier}
\label{sec:proposed}

\subsection{System Model}
Following~\cite{Wang2010}, we assume a sequence of $M$ discrete, complex, i.i.d. and sampled baseband symbols, $\mathbf{s}^{(k)}\triangleq[s_1^{(k)} \cdots  s_M^{(k)}]$, drawn from a constellation $\mathcal{M}_k\in\left\{\mathcal{M}_1,\ldots,\mathcal{M}_K\right\}$, transmitted over AWGN channel. The received signal, under constellation $\mathcal{M}_k$, is given as $\mathbf{r}\triangleq[r_1 \cdots r_M]$, where $r_n=s_n^{(k)}+g_n$, $g_n\sim\mathcal{CN}\left(0,\sigma^{2}\right)$. We further define the SNR as $E[(s_n^{(k)})^2]/\sigma^2$. The task of the modulation classifier is to find $\mathcal{M}_{\hat{k}}$, from which $\mathbf{r}$ is drawn from. Without loss of generality, we consider unit power constellations.

\subsection{Classification Based on Sampled Distribution Distance}
Let $\mathbf{z}\triangleq[z_1 \cdots z_N]=f(\mathbf{r})$ where $f(\cdot)$ is the chosen mapping from received symbols $\mathbf{r}$ to the extracted feature vector $\mathbf{z}$, where $N$ is the length of the feature vector. Possible feature maps include $|\mathbf{r}|$ (magnitude, $N=M$), the concatenation of $\Re\{\mathbf{r}\}$ and $\Im\{\mathbf{r}\}$ (quadrature, $N=2M$), the phase information $\angle{\mathbf{r}}$ (angle, $N=M$), among others. The theoretical CDF of $z_i$ given $\mathcal{M}_k$ and $\sigma^2$, denoted as $F_0^k(z)$, is assumed to be known \emph{a priori} (methods of obtaining these distributions, both empirically and theoretically, are presented in~\cite[Sec. III-A]{Wang2010}). 

In this paper we focus on algorithms that use the ECDF defined as
\begin{equation}
\label{eq:ecdf}
F_{N}(t)=\frac{1}{N}\sum\limits_{n=1}^{N}\mathcal{\mathbb{I}}(z_n\leq t),
\end{equation}
as the discriminating feature for classification. Here,  $\mathbb{I}(\cdot)$ is the indicator function whose value is 1 if the function argument is true, and 0 otherwise. If the complete ECDF resulting from the entire feature vector, $\mathbf{z}$, is used for classification, we get the conventional distribution distance measures such as Kuiper, Kolmogorov-Smirnov, Anderson-Darling and others. Details of these measures are discussed in \cite{Stephens1974}. Once the ECDF is found and the appropriate distribution distance is calculated, the candidate constellation with minimum distance is chosen. However, prior work in \cite{Urriza2011,Wang2012} have shown that improved classification accuracy can be achieved at much lower computational complexity and with increased model robustness by finding the value of the ECDF at a small number of specific testpoints. 

We describe these methods formally by defining a set of $L$ testpoints: $\mathbf{t}=[t_1 \cdots t_L]$, with $t_{i+1} \geq t_i$. For notational consistency, we also define the following virtual test points, $t_0\triangleq-\infty$ and $t_{L+1}\triangleq+\infty$ in addition to $\mathbf{t}$.  Evaluating the ECDF from (\ref{eq:ecdf}) at $\mathbf{t}$ gives us $\mathbf{x}=[x_1 \cdots x_L]$, $x_i\triangleq F_N(t_{i})$. We refer to any classifier that utilizes the feature vector $\mathbf{x}$ as a \emph{sampled distribution distance-based classifier}. As an example, the variational distance (VD) classifier from \cite{Wang2012} proposed forming $\mathbf{t}$ from ECDF points that give either a local maxima or minima of the difference between two theoretical cdfs of the candidate classes. Instead of using the sampled ECDF directly, VD classifier finds the number of samples that fall between two consecutive testpoints, which is equivalent to taking the difference of the ECDF at consecutive testpoints, $F_N(t_{i}) - F_N(t_{i-1})$.

In this paper our goal is to optimize the classification accuracy of the sampled distribution distance classification approach defined as
\begin{equation}
P_C=\sum_{i=1}^K\Pr(\hat{k}=i\left|\mathcal{M}_i\right.)\Pr\left(\mathcal{M}_i\right).
\end{equation}
Intuitively, there are two ways to improve $P_C$. First, since different testpoints have varying distribution distance, it is expected that different weights should be assigned to each testpoint. Second, the choice of the number and location of the points along the ECDF should also be investigated to find the proper balance between complexity and classification accuracy. Both of these improvements are addressed in the following subsection.

\subsection{Proposed Classifier}
\label{subsec:optimal}
We first assume that $\mathbf{t}$ has been selected \textit{a priori} and our goal is to find the optimal classifier for the resulting feature vector $\mathbf{x}$. We want to find a discriminant function $g_k(\mathbf{x})$ for each $k\in[1,K]$, for every candidate constellation $\mathcal{M}_k$. Where we follow the rule:
\begin{equation}
\label{eq:rule}
\text{Choose: }\mathcal{M}_i \text{ s.t. } g_i(\mathbf{x}) > g_j(\mathbf{x})\, \forall\, j\neq i
\end{equation}

It is well established in decision theory that if the performance metric used is average classification error, the optimal classifier is based on the \emph{Bayes decision procedure} \cite{Duda2001}. This procedure can be stated as:
\begin{equation}
\text{Choose: }\mathcal{M}_{i}\text{ s.t. }\Pr(\mathcal{M}_{i}\left|\mathbf{x}\right.)>\Pr(\mathcal{M}_{j}\left|\mathbf{x}\right.)\,\forall\, j\neq i.
\label{eq:bdp}
\end{equation}

Using the prior probabilities $\Pr(\mathcal{M}_{i})$, the posterior probabilities $\Pr(\mathcal{M}_{i}\left|\mathbf{x}\right.)$ could be found from $\Pr(\mathbf{x}\left|\mathcal{M}_{i}\right.)$ using Bayes formula.
Thus, finding the pdf of the feature vector conditioned on the modulation scheme, $\Pr\left(\mathbf{x}\left|\mathcal{M}_{i}\right.\right)$, effectively gives us the optimal classifier in the minimum error rate sense.

The testpoints partition $\mathbf{z}$ into $L+1$ regions. An individual sample, $z_n$, can be in region $l$, such that $t_{l-1} < z_n \leq t_{l}$, with a given probability, completely determined by the cdf, $F_0^k(z)$. The number of samples that fall into each of the regions, $\mathbf{n}\triangleq [n_1 \cdots n_{L+1}]$, where $n_i$ corresponds to region $i$, $1\leq i \leq L+1$,  is jointly distributed according to a multinomial probability mass function (pmf) given as
\begin{equation}
f(\mathbf{n} | N,\mathbf{p}) = 
\begin{cases}
\frac{N! p_1^{n_1} \cdots p_{L+1}^{n_{L+1}}}{{n_1}! \cdots {n_{L+1}}!},& \text{if $\sum\limits_{i=1}^{{L+1}}n_i=N$},\\
0,& \text{otherwise},
\end{cases}
\label{eq:multinomial}
\end{equation}
where $\mathbf{p}\triangleq [p_1 \cdots p_{L+1}]$, and $p_l$ is the probability of an individual sample being in region $l$. Given that $\mathbf{z}$ is drawn from $\mathcal{M}_k$, $p_l=F_0^k(t_l)-F_0^k(t_{l-1})$, for $0<l\leq L+1$.

Given a particular $\mathbf{x}$, the number of samples in each of the $L+1$ regions could be found as $n_{i}=N\left(x_{i}-x_{i-1}\right)$ where $x_0\triangleq 0$ and $x_{L+1}\triangleq 1$. This gives a mapping from any given $\mathbf{x}$ to $\mathbf{n}$ and therefore to the pmf $f(\mathbf{n} | N,\mathbf{p})$ as defined in (\ref{eq:multinomial}). Therefore we have the complete class-conditional pdf, $\Pr\left(\mathbf{x}\left|\mathcal{M}_{k}\right.\right)$ with $\mathbf{p}$ in (\ref{eq:multinomial}) determined by $F_0^k(z)$, the cdf of class $\mathcal{M}_k$. Thus we have the optimal classifier. We will refer to $\mathbf{x}$ and $\mathbf{n}$ conditioned on class $\mathcal{M}_k$ as $\mathbf{x}^{(k)}$ and $\mathbf{n}^{(k)}$.

Although the multinomial pmf in (\ref{eq:multinomial}) can be used for minimum error rate classification, its calculation is very computationally intensive. To address this issue we note that asymptotically the multinomial pmf, $f(\mathbf{n} | N,\mathbf{p})$ in (\ref{eq:multinomial}), approaches a multivariate Gaussian distribution, $\mathbf{n}^{(k)}\sim\mathcal{N}(\boldsymbol{\mu}_k^{(n)},\mathbf{\Sigma}_k^{(n)})$ as $N\rightarrow\infty$. Where,
\begin{eqnarray}
\boldsymbol{\mu}_k^{(n)}&=&N\mathbf{p}\\
\{\mathbf{\Sigma}_k^{(n)}\}_{ij}&=&\begin{cases}
Np_i(1-p_i), & \text{if}\,i=j,\\
-Np_ip_j, & \text{if}\,i\neq j.
\end{cases}
\end{eqnarray}
Since $\mathbf{x}$ is simply the cumulative sum of $\mathbf{n}$ (i.e. $x_i=\sum_{j=1}^i n_j$), which is a linear operation, it follows that $\mathbf{x}^{(k)}\sim\mathcal{N}(\boldsymbol{\mu}_k,\mathbf{\Sigma}_k)$ where,
\begin{eqnarray}
\label{eq:mu}
\{\boldsymbol{\mu}_k\}_i&=&N\sum_{j=1}^i p_j=F_0^k(t_i),\\
\label{eq:sigma}
\{\mathbf{\Sigma}_k\}_{ij}&=&\sum_{l=1}^i\sum_{m=1}^j \{\mathbf{\Sigma}_k^{(n)}\}_{lm}.
\end{eqnarray}

Having shown that the feature vector $\mathbf{x}$ is asymptotically Gaussian distributed, we can proceed to apply the \emph{Bayes decision procedure} in (\ref{eq:bdp}). However, the full multivariate pdfs are not required to perform classification because the optimal discriminant functions for Gaussian feature vectors are known to be quadratic with the following form \cite{Duda2001}:
\begin{equation}
\label{eq:discriminant}
g_{k}(\mathbf{x})=\mathbf{x}^{T}\mathbf{W}_{k}\mathbf{x}+\mathbf{w}_{k}^{T}\mathbf{x}+w_{k0},
\end{equation}
where
\begin{equation}
\mathbf{W}_{k}=-\frac{1}{2}\mathbf{\Sigma}_{k}^{-1},\:\mathbf{w}_{k}=\mathbf{\Sigma}_{k}^{-1}\boldsymbol{\mu}_{k},
\end{equation}
and
\begin{equation}
w_{k0}=-\frac{1}{2}\boldsymbol{\mu}_{k}^{T}\mathbf{\Sigma}_{k}^{-1}\boldsymbol{\mu}_{k}-\frac{1}{2}\ln\left|\mathbf{\Sigma}_{k}\right|+\ln\Pr\left(\mathcal{M}_{k}\right).
\end{equation}
In the following sections we will simply refer to this classifier as the Bayesian approach.

\subsection{Note on Implementation}
Similar to rcK \cite{Urriza2011} and VD \cite{Wang2012} the Bayesian approach only needs to store the testpoint locations for a fixed set of SNRs since the theoretical cdf is dependent on SNR. Given a $\mathbf{t}$ of size $L$, VD and rcK require both $\mathbf{t}$ and $\boldsymbol{\mu}_k$ for each class $\mathcal{M}_k$. In contrast, the Bayesian approach requires the same vector $\mathbf{t}$, an $L\times L$ matrix $\mathbf{W}_k$, a vector $\mathbf{w}_k$ of size $L$, and a scalar $w_{k0}$ for each class $\mathcal{M}_k$. However, there are typically no more than 12 testpoints (total number of pdf-crossings), so this additional storage requirements are negligible. The Bayesian approach also requires the calculation of a quadratic form expression (\ref{eq:discriminant}). Again, due to the fact that only a relatively small number of testpoints is used, the additional complexity is minimal. 

\subsection{Testpoint Selection}
In this subsection we present a method for choosing testpoint locations, $\mathbf{t}$, that provide good classification performance. The method of using the pdf-crossings make intuitive sense, since it tries to find the testpoints that provide the maximum difference in the theoretical cdf while providing some heuristic rule that the testpoints will be spaced apart. Tespoints that are too close to each other are not as effective because the ECDF tends to be highly correlated and thus provide minimal additional information.

Another issue with using the pdf-crossing is that it does not factor in knowledge of the correlation between testpoints. As we have shown in Section~\ref{subsec:optimal}, the distribution $\mathbf{x}$ follows an approximate multivariate Gaussian with statistics given in (\ref{eq:mu}) and (\ref{eq:sigma}). Therefore, the class-conditional means $\boldsymbol{\mu}_k$  and covariance matrices $\mathbf{\Sigma}_k$ are sufficient to completely describe the distribution of the feature vectors conditioned on $\mathcal{M}_k$. Thus, these statistics are also sufficient to find the optimal testpoint locations, $\mathbf{t}^*$.

However, since $\mathbf{\Sigma}_k$ are clearly not  equal for all $\mathcal{M}_k$, a closed form expression for the classification accuracy for this problem does not exist. Instead, a $K$-dimensional integration is required and the limits, determined by the decision boundaries defined by (\ref{eq:discriminant}), are non-trivial. As is typically done in this scenario, we replace exact $P_C$ with a sub-optimum distance metric that is easier to evaluate and does not require a $K$-dimensional integral. In particular we use the Bhattacharyya distance first studied for signal selection in \cite{Kailath1967} shown to be a very effective as a ``goodness'' criterion in the process of of selecting effective features to be used in classification. The metric is shown here for reference:
\begin{eqnarray}
D_{B}=\frac{1}{8}\left(\boldsymbol{\mu}_{1}-\boldsymbol{\mu}_{2}\right)^{T}\mathbf{\Sigma}^{-1}\left(\boldsymbol{\mu}_{1}-\boldsymbol{\mu}_{2}\right)\nonumber\\
+\frac{1}{2}\ln\left(\frac{\left\vert\left(\mathbf{\Sigma}_{1} +\mathbf{\Sigma}_{2}\right)/2\right\vert}{\sqrt{\vert\mathbf{\Sigma}_{1}\vert\vert\mathbf{\Sigma}_{2}\vert}}\right).
\end{eqnarray}

Note that the Bhattacharyya distance is calculated between 2 classes. As a result, the search for testpoints can only be performed for the $K=2$ case. However, this could be done sequentially through all the possible pairs of $\mathcal{M}_k$. As $D_B$ is a function of $\boldsymbol{\mu}_k$ and $\mathbf{\Sigma}_k$ which are functions of our testpoint selection, $\mathbf{t}$, then we can express it as $D_B(\mathbf{t})$. We thus find the good candidate testpoint by
\begin{equation}
\mathbf{t}^{*}=\arg\max_{\mathbf{t}}D_{B}\left(\mathbf{t}\right),
\end{equation}
under the constraint $t_{i+1} \geq t_i$.

As this is an $L$-dimensional optimization problem, a closed-form solution is beyond the scope of this letter paper. Instead, we turn to numerical optimization methods (gradient descent methods) to find local maxima. The intial point of these procedures could be chosen to coincide with the pdf-crossings or equally spaced over some interval.


\section{Results and Discussion}
\label{sec:results}

\subsection{Testpoint Selection}
\begin{figure}
\begin{center}
\includegraphics[width=\columnwidth]{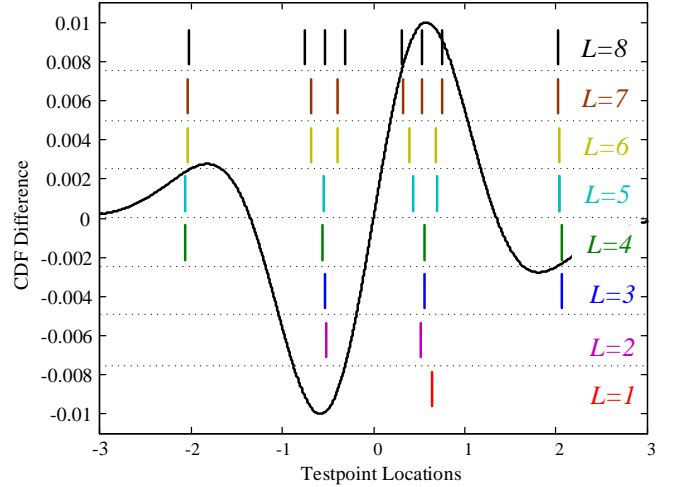}
\end{center}
\caption{Optimized testpoint locations for varying number of testpoints, $L$. The solid line shows the CDF difference between the two classes (4-QAM and 16-QAM, under SNR=0 dB, $M=200$)}
\label{fig:testpoints}
\end{figure}

For the results section we focus on the quadrature feature which is a concatenation of the I and Q component of each symbol. In Fig.~\ref{fig:testpoints}, we show the results of the testpoint selection procedure with $N=200$, under 0 dB SNR, for varying number of testpoints with the two class being 4-QAM and 16-QAM.
The solid line plot corresponds to the difference of the two theoretical CDFs. We note that in the VD classifier the local maxima and minima of this plot are used as the testpoints. However, we find that the numerical optimization finds ``good'' testpoints to be close, but not exactly at the local maxima and minima. This is due to the additional information provided by the covariance matrices.

In contrast to VD classifier that has a fixed number of testpoints (4 for this particular problem) corresponding to the number of local maxima and minima, the optimization procedure allows more flexibility in choosing the number of testpoints. In Fig.~\ref{fig:testpoints}, we show the result of the optimization procedure for a range of 1 to 8 testpoints. It confirms our intuition that ``good'' testpoints tend to be 1) spaced apart to avoid high correlation, 2) concentrated around locations that have high CDF difference, and 3) are not necessarily the same for different values of $L$. This result further confirms the need to jointly optimize the testpoint locations.

\subsection{Comparison With Existing Techniques}
\begin{figure}
\begin{center}
\includegraphics[width=\columnwidth]{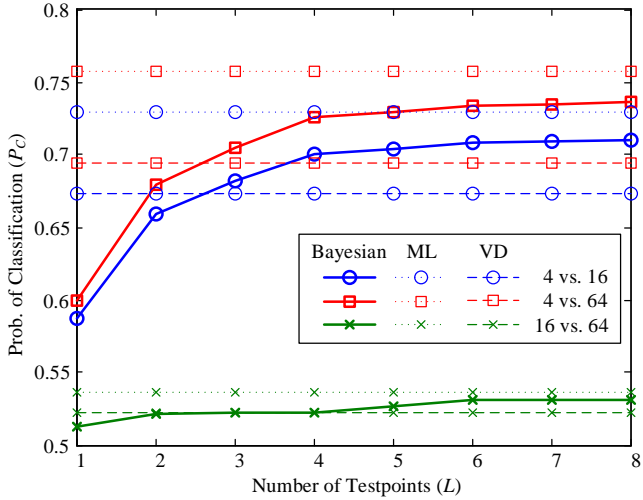}
\end{center}
\vspace{-3mm}
\caption{Effect of increasing number of testpoints on $P_C$ for all possible pairs of constellations of interest.The classification accuracy of both ML and VD classifiers are also shown for comparison. (SNR=0 dB, $M$=200)}
\vspace{-2mm}
\label{fig:vary_tp}
\end{figure}

As mentioned in the previous section, the proposed approach has the flexibility of varying the number of testpoints. This effectively gives more flexibility to trade-off classification accuracy with computational complexity. This idea is illustrated in Fig.~\ref{fig:vary_tp}. For $N=200$ and SNR=0 dB, we show the classification accuracy of the proposed method as the number of testpoints is increased from 1 to 8, for all possible pairs of $\mathcal{M}_k$. The dotted lines correspond to the accuracy of the ML classifier which serves as an upperbound to classification accuracy, while the dashed lines correspond to that of the VD classifier. Note that both are plotted as horizontal lines because ML does not utilize testpoints, while VD has a fixed number of testpoints corresponding to the pdf-crossings.

We see that the proposed method is able to exceed the accuracy of the VD classifier with as low as 3 testpoints. Further, the method's accuracy could be improved by adding more testpoints but at the cost of higher complexity. We also note that with additional testpoints, the Bayesian classifier reaches classification accuracy close to the ML classifier.

\begin{figure}
\begin{center}
\includegraphics[width=\columnwidth]{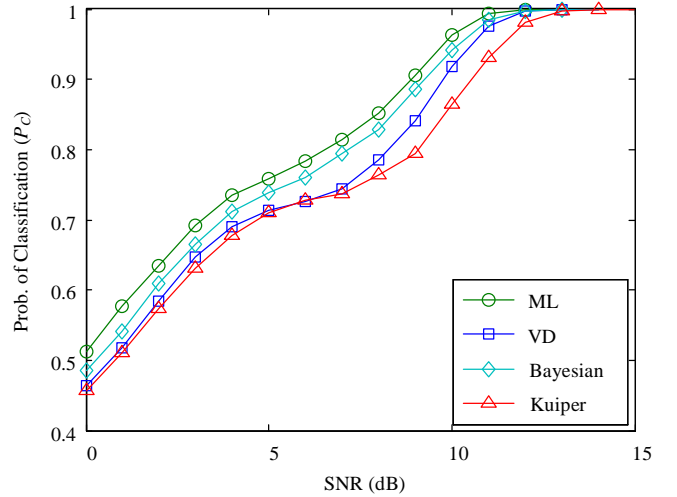}
\end{center}
\vspace{-3mm}
\caption{Comparison of the proposed Bayesian method with other existing approaches under varying SNR with $M$=200 symbols used for classification. The same number of testpoints are used for both VD and Bayesian.}\vspace{-2mm}
\label{fig:vary_snr}
\end{figure}

Finally, in Fig.~\ref{fig:vary_snr}, we compare the performance of the proposed method with the existing techniques under varying SNR with $M=200$ symbols used for classification. To have a fair comparison, the same number of testpoints are used for both VD and Bayesian. For the entire range of SNR the proposed Bayesian approach is shown to provide substantial gains over the VD classifier. We emphasize again that asymptotically, the proposed approach is the optimal classifier when using the sampled distribution distance as the discriminating feature. Also shown in the plot are the classification accuracy of the ML classifier which acts as the upperbound, and the conventional Kuiper classifier.

\section{Conclusion}
\label{sec:conclusion}
In this letter we presented the optimal discriminant functions for classifying using the sampled distribution distance. This method was shown to provide substantial gains compared to other existing approaches. The performance of this method is also shown to be close to the ML classifier but at significantly lower computational complexity. Although modulation classification is presented in this paper to illustrate the basic concept, the approach is not limited to this application. The same classifier can be generalized to any classification problem where the cdf of each class is available.

\end{document}